\def\year{2016}\relax
\documentclass[letterpaper]{article}
\usepackage{aaai16}
\usepackage{times}
\usepackage{helvet}
\usepackage{courier}
\usepackage{microtype}
\usepackage{csquotes}
\usepackage{graphicx}
\begin{document}
%
\title{Off The Beaten Lane:\\AI Challenges In MOBAs Beyond Player Control}
\author{
Michael Cook\\
Games Academy\\
Falmouth University\\
\And
Adam Summerville\\
Expressive Intelligence Studio\\
UC Santa Cruz\\
\And
Simon Colton\\
Games Academy\\
Falmouth University\\}
\maketitle
\begin{abstract}
MOBAs represent a huge segment of online gaming and are growing as both an eSport and a casual genre. The natural starting point for AI researchers interested in MOBAs is to develop an AI to play the game better than a human - but MOBAs have many more challenges besides adversarial AI. In this paper we introduce the reader to the wider context of MOBA culture, propose a range of challenges faced by the community today, and posit concrete AI projects that can be undertaken to begin solving them.
\end{abstract}

\section{Introduction}
Multiplayer Online Battle Arenas, a clumsy phrase shortened to MOBA, describes a growing genre of videogames typically designed as highly competitive eSports. Many of them are part of a family tree that traces back to Warcraft 3 mods such as \textit{DOTA Allstars}, and now form a contingent of the most popular and most player games of today, including \textit{League of Legends}, \textit{DOTA 2} and \textit{Heroes Of The Storm}.

MOBAs are huge and rapidly expanding in every sense - in terms of their cultural impact on games, in terms of their financial impact on the industry, and in terms of their impact on how the games community is spending its time. As a result, they are attracting increasing interest from researchers, both those interested in building technological systems into the game, and those interested in studying the many different stakeholder groups in the community. 

In an age of deep learning and AIs making headlines, the obvious target for AI researchers to study would be the implementation of bots that can play the game at human or superhuman level. Google DeepMind have already stated that their next objective will be to develop a system that can play Starcraft 2, a competitive videogame with similarities to MOBAs at least culturally if not mechanically. We believe that the development of player-competitive AI represents just one small challenge offered by the complex world of MOBA playing, but that many of these challenges are hard for researchers to access because of the vast knowledge barriers that exist when trying to understand the genre or what makes it interesting.

In this paper we try to pick apart interesting challenges in and around the MOBA genre. Some of them touch upon deeply technical problems within the game, others  relate to ways that AI can assist in the emerging activities that happen around MOBA games, especially relating to the professional eSports scene. We do our best to present these challenges with the minimum of context for the reader, so that researchers whose background may not be in strategy games or MOBAs specifically can hopefully see why these problems are interesting and hopefully motivate more diverse research in the area.

The remainder of the paper is organised as follows: in \textit{Background} we provide a minimal introduction to MOBAs, primarily focusing on one specific game, DOTA 2 due to the particular experience of the authors. We discuss the game itself, the professional scene, and surrounding culture. In \textit{Challenges} we outline some important challenges we have identified throughout the MOBA genre, and provide specific ideas about projects that could be conducted to help work towards solving or assisting humans in tackling these problems. In Existing \& Related Work we discuss some of the work already done in the MOBA genre, as well as highlighting some work beyond MOBAs that could contribute to the challenges in this paper.

\section{Background - DOTA 2}
DOTA 2 is a complicated game with a large amount of intersecting systems, lists of specific knowledge and historical trends. For the purposes of this paper we will be focusing on a light coverage of key themes and ideas, sufficient to understand the challenges we present in the next section. Additionally, although DOTA 2 is a game about exceptions and special cases, we will make generalisations in this section in order to simplify the high-level description of the game.

\subsection{Gameplay - Drafting}
\begin{figure}
\includegraphics[width=\columnwidth]{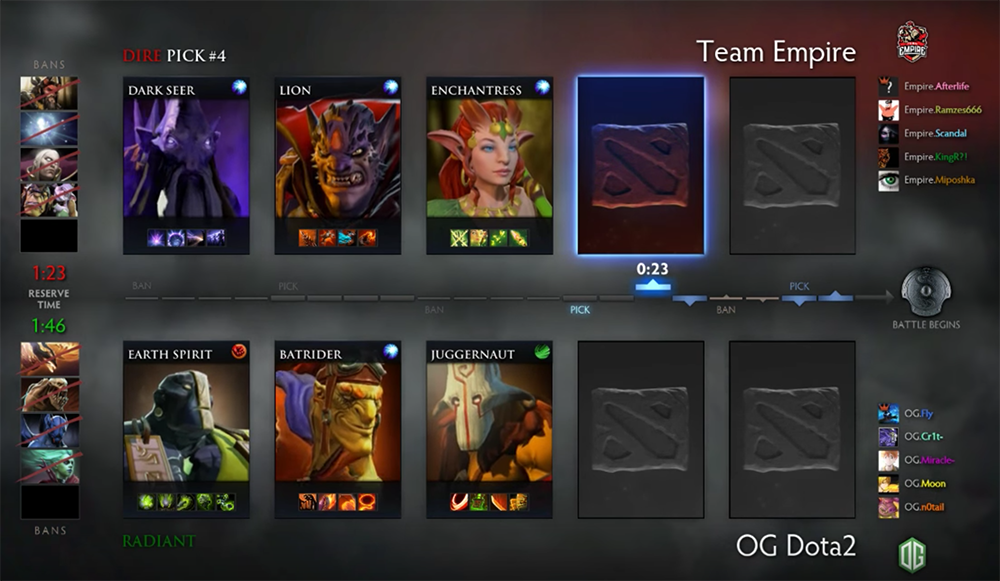}
\caption{The drafting phase from a match between Team Empire and OG. Picked heroes appear in large portraits, while bans are listed on the left in smaller images. Team Empire are about to pick their fourth hero.}
\label{fig:draft}
\end{figure}

A single game typically has two phases -- a \textit{drafting phase} where players choose heroes to play, and a \textit{gameplay phase} where the main game takes place. The nature of the draft phase depends on the game mode. In this paper we will primarily discuss \textit{Captain's Mode}, a game mode used in professional play but less popular with casual players. In Captain's Mode each player nominates a captain who chooses heroes for their team. The two captains go through a fixed order of \textit{picking} heroes (which adds them to their team) and \textit{banning} heroes (which stops either team from picking them). A hero can only be picked once. Figure \ref{fig:draft} shows a screenshot from a professional game's drafting phase. Drafting is complex enough to warrant its own paper. Indeed, professional players often turn up to tournaments with `drafting bibles' filled with pages of notes about enemy team strategies, their own prepared ideas for drafts, and historical information about synergies and counter-strategies. 

There are 112 heroes in the game in total, and each one has a particular combination of skills and strengths, some of which may overlap or synergise with other heroes. For the purposes of this paper, we will simplify heroes into two kinds: \textit{support} heroes and \textit{core} heroes. Support heroes are strong in the early parts of a game, typically because they have damaging spells or spells with special effects (such as a \textit{stun}, which temporarily stops a hero from acting). Core heroes have properties which make them better later in the game, often because they either need to buy a particular item to become effective, or because they scale up faster than other heroes. For example, one hero is able to attack up to five targets simultaneously. This means that an item which provides increased damage is potentially five times as valuable on that hero. The number of cores selected for a team, and how much time they need to spend collecting resources, affects the kind of strategies available to a team during the game itself.

\subsection{Gameplay - Match} 
\begin{figure}
\includegraphics[width=\columnwidth]{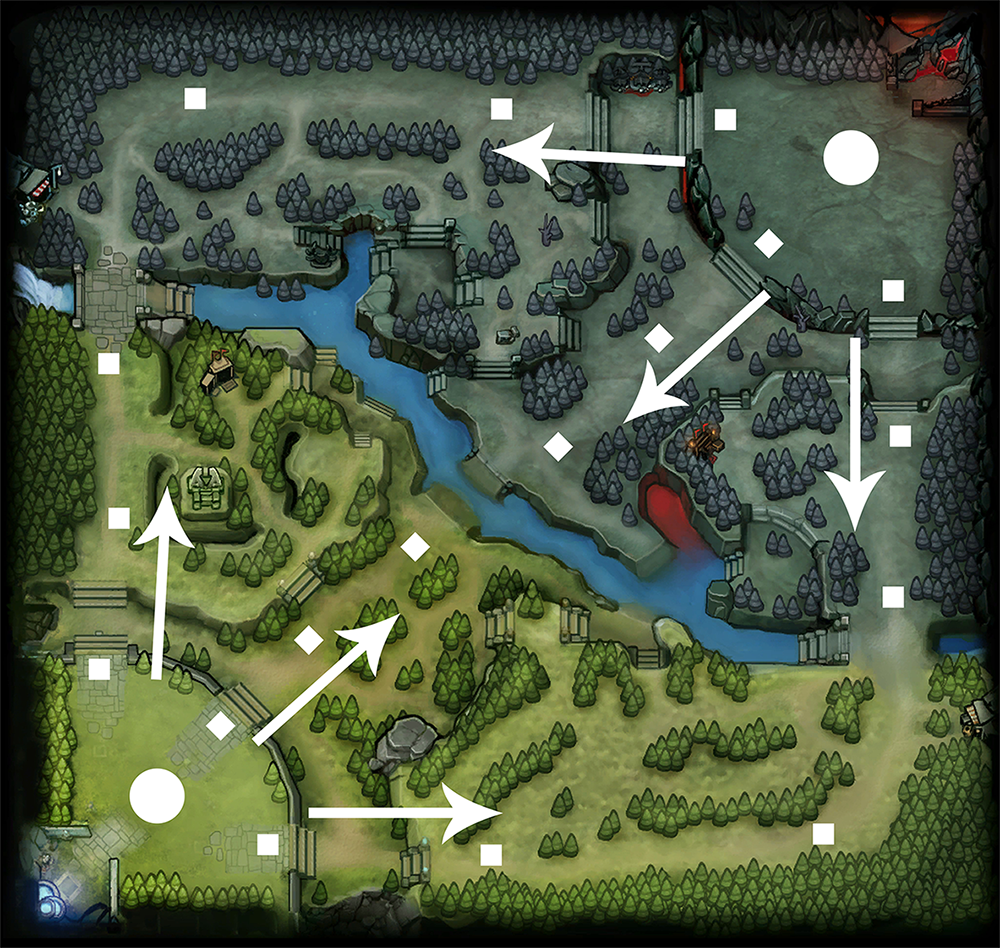}
\caption{An annotated map from a version of DOTA 2. Arrows indicate the directions creeps move along the three lanes. Squares indicate the location of towers, while the large circle is each team's Ancient.}
\label{fig:map}
\end{figure}

The primary objective in a game of DOTA 2 is to destroy the opposing team's Ancient, a structure protected in the centre of each team's base. Figure \ref{fig:map} shows an annotated overheap drawing of the entire DOTA 2 map\footnote{This map is from an earlier version of the game but is used here as the key details remain the same.} with the Ancients represented as circles in opposite corners. Connecting the two bases are three pathways called lanes, two around the edges of the map and one through the centre. Periodically, NPC creatures called creeps spawn from both bases and travel along these lanes, attacking each other when they meet. 

Towers are also placed along these lanes, indicated by squares in Figure \ref{fig:map}. Towers are strong structures which attack nearby enemy units, provide vision for their team, and help define the borders of a team's territory. Towers are invulnerable to damage unless they are the outermost tower in a lane, and the Ancient is invulnerable to damage unless one of the three towers at the end of a lane are destroyed. Thus, destroying towers in lanes is an important marker of overall progress in the game, and many strategies exist to achieve this, including some which intentionally avoid fighting enemy heroes in favour of destroying buildings.

Games are colloquially broken up into three phases. The first phase is the \textit{early game}, in which core players are focusing on obtaining gold and experience points. During this phase support players may protect their team's cores, or they may attempt to move around the map and attack the opposing team's cores. The second phase is the \textit{mid-game}, where some cores have reached a point where they can begin to dictate the game's direction. This might involve grouping with support heroes to destroy towers, or moving as a larger group to fight enemy heroes. Some cores may continue to obtain gold and experience if they need more than other cores. The final phase is the \textit{late game}. At this point heroes have generally stopped focusing on resources and have important items and skills. This phase of the game is much less predictable as players tend to have less experience of it (all matches have an early game, but only some last long enough to enter the late game). This phase involves a larger amount of movement, fighting and strategic decision-making. 

\subsection{Professional Play}
In 2011 Valve announced DOTA 2 by organising an global tournament called The International, in which sixteen professional DOTA 1 teams were invited to compete for a share of \$1.6m, the largest prize pool in eSports at the time. In 2015 The International's fifth incarnation had a prize pool of over \$18m. In the intervening years the professional scene has grown to become both an aspiration for young players and a big business for sponsors and organisers. Most of the top teams are run by independent organisations, have multiple sponsors, and employ managers, coaches, dedicated analysts, PR and various other staff.

Most DOTA 2 tournaments, from the smallest amateur regionals to the highest-tier multi-million viewer events, are broadcast online to watch for free, with \textit{casters} providing commentary, analysis and statistics on the matches. This is common across most eSports, including \textit{Heartstone}, \textit{Starcraft 2} and other MOBAs like \textit{League of Legends}. A common structure for an eSports broadcast has two distinct panels of people: an \textit{analysis desk} which discusses games during downtime between matches; and a \textit{caster desk} which describes and comments on the action during a game. Larger broadcasts may also employ \textit{statistics analysts} who listen to commentary and provide relevant statistics about past events or current trends, as well as \textit{dedicated observers} responsible for controlling the in-game camera based on commentator interest.

\section{Challenges}
In this section we describe challenges besides AI character control that we believe will become important and fruitful areas of research in the next few years. In each case we provide a little additional background where necessary, state the nature of the problem and how it is currently tackled (if at all) and then propose possible routes for AI research to take.

\subsection{Commentary}
Commentary is broken into three distinct areas, each with their own unique challenges - draft analysis, play-by-play and hypecasting.

\subsubsection{Draft Analysis}
During the draft the analysis desk typically discusses the heroes being picked and banned, the wider context of the game and how it reflects the current metagame, and how they expect the game itself to play out when the draft is complete. A draft timer limits the amount of time teams can spend picking heroes, but teams tend to maximise their use of this time, meaning drafts are often slow-paced and thus are a good area for researchers to focus on initially for commentary generation and assistance.

Good predictive models for which heroes are likely to be picked or banned are fundamental to AI draft analysis. Machine learning is likely to perform well here, but the problem is complicated by the relative lack of data on professional matches. Towards the end of a particular patch cycle (discussed later) there may be a few thousand matches to pull from at most. More specific predictions may have much smaller datasets -- data on a specific team, for example, may include only a few dozen matches. Hybrid techniques may be needed to provide useful analyses at different levels of detail, employing decision trees or linear regressions for smaller data pools.

A secondary problem for good draft analysis, which also applies to statistics provided during the game, is selection. A vast quantity of data is available for DOTA 2, from the heroes drafted to the exact time at which a player bought an item in a specific game. Facts of all shapes and sizes can be discovered -- records set by players, trends or preferences of certain captains, historical stories and expected performances -- but selecting which are most appropriate for the current discussion is a separate challenge. This is partly a natural language processing problem, monitoring the conversation between analysts to dynamically rank facts and statistics for relevance, but models of `interestingness' are also important. Subjective interestingness, and notions of `actionability' and `unexpectedness' from the knowledge discovery domain \cite{kd}, may help inform models for statistic selection.

\subsubsection{Play-By-Play}
Commentating a game involves solving several different problems simultaneously and in real time. First and foremost, commentators must identify the most salient events happening in the game and describe them succinctly. During moments where many heroes are fighting together the challenge comes from identifying the most important features of the fight -- which player actions are having the most impact, which actions are causing important changes in who has advantage. A major challenge here is ranking the actions of ten players simultaneously and being able to predict the next few seconds of action (since it takes time to commentate on an action, during which time more events have taken place). 

When large fights are not happening, the choice of what to discuss becomes broader. Commentators may talk about the emerging themes in the game -- what strategies the teams are employing, how those strategies are faring, or the individual performances of a particular player. They may also identify what teams are planning to do next, based on the behaviour of players currently. In periods of downtime commentators may also discuss the game in the wider context of the tournament or series it is in -- how it reflects the trends elsewhere in the tournament, or how the winner of this match might fare against the other teams they are likely to face next.

\subsubsection{Hypecasting \& Wordplay}
It's tempting to consider commentary as being purely concerned with conveying the facts of the game to the viewer, but a major part of MOBA commentary is in providing a sense of playfulness and excitement to proceedings. For some casters this is achieved through a sense of power and intensity in their delivery -- Tobi Dawson is famous for his high-energy commentary and emotive descriptions of play. Other casters employ colourful language, phrasing and inventive wordplay to entertain viewers. Below is an excerpt from the grand final of The International in 2012, where David Gorman is addressing his co-caster, David `Lumi' Zhang. 

\begin{displayquote}
\textit{There's a freight train running down the tracks and it's about to hit a car, and let me tell you Lumi, the car gives way -- not the freight train.}
\end{displayquote}

Inventing clever descriptions for events and coining names for teams or players links in well with existing research into metaphor, analogy and humour in computational creativity. Work in \cite{ritchie} lays out ways to create humour from linguistic collisions of concepts, and \cite{veale} describes systems for making perceptual connections between the real world and linguistic constructions. This shows that we can do more with language than simply state what is happening on screen, and research into AI casters should endeavour to go beyond this and be as engaging and innovative with language as human casters are.

%

\subsection{Camera Control}
\subsubsection{Automated Camera Control}
In the early days of eSports commentators would broadcast their screen directly to viewers, and simultaneously describe what was happening as they tried to smoothly direct the camera's view of proceedings. Today, the bigger tournaments and studios hire dedicated observers whose only job is to control the camera and focus on the most relevant action. Camera control is much harder in eSports compared to traditional sports as games often lack a single point of focus (such as a ball). Deciding what to show and how to show it is a difficult task. Tools that can help automate or assist in this process not only help high-end broadcasters, but also help amateur broadcasters and casual spectators a chance to focus on the game itself rather than controlling a camera.

There is also a healthy body of research relating to camera control for other game genres -- in \cite{affective} the authors specifically consider the affective qualities of automated camera control, which is highly relevant for framing different kinds of action in a hectic MOBA game, while in \cite{dynamic} the authors look at camera control as a multi-objective optimisation problem which is particularly appropriate for the dynamic action in spectator eSports.

\subsubsection{Highlighting}
An emerging problem for eSports design in general is a desire for succinct highlighting of matches. This is something that game designers are showing interest in not only at the professional level, highlighting important moments from top matches, but at the casual level where individual games may be able to produce their own highlight reels. DOTA 2 can be asked to automatically provide highlights of any game replay file, although the process is rudimentary and has not been improved in years. 

Highlighting is a mix of problems from camera control and commentary, with an additional pressure to present the results in an engaging and exciting way. The most interesting and significant moments from a game must be extracted, and framed in a way that best displays the action at hand. While many AI techniques could be used here, one promising possibility would be to leverage Twitch chat as a form of supervised learning. Twitch, a popular streaming service used to broadcast most professional DOTA 2 live, has a prolific (and, we should stress, often toxic) chat community. At times of intense action, humour, surprise or skill chatters often simultaneously send certain emotes in large quantities. These emote surges can be used to automatically label events in certain professional replays\footnote{Inspired by http://www.skip2.tv which implemented an emote counter to identify exciting moments in Twitch streams}, which could then be cross-referenced with replay data to train systems to recognise what causes such a strong reaction. These models could then be reapplied to ordinary games as an automated highlighter.

\subsection{Vision}
An \textit{observer ward} is a cheap but limited-quantity item that can be placed on the map to provide vision in a 360 circle, obscured by certain map features like trees or cliffs. Observer wards are invisible, but can be detected and destroyed by certain items. If not destroyed, they expire naturally after seven minutes. Because observer wards are limited in number but destroyable, a tension exists between placing wards in good places while avoiding becoming predictable. Professional teams often study how certain players place wards in order to predict their behaviour in future games.

\begin{figure}
\includegraphics[width=\columnwidth]{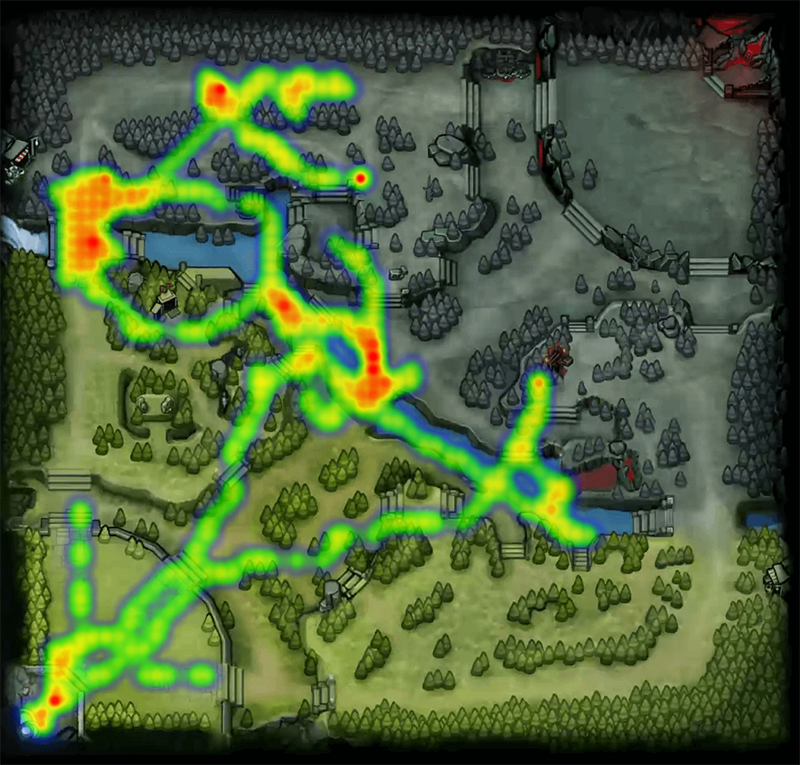}
\caption{A heatmap for a single player's movement during the first ten minutes of a DOTA 2 match.}
\label{heatmap}
\end{figure}

This is an example of a tightly defined system within DOTA 2 that is ripe for analysis and optimisation. Publicly available replay data for any DOTA 2 match, including professional games, includes data about hero movement and ward placement, allowing a complete model to be built for which players were visible at what times, and what wards are commonly placed. Figure \ref{heatmap} shows a heatmap of a single player's movement in the early game phase of a match. A tool to analyse the warding habits of a team to predict placement in an upcoming game, or to analyse common player movements and suggest ward placements for optimal coverage, would be useful for all players from new learners to professional teams. 

Existing research into modelling player vision may inform further work here. For example, in \cite{tremblay} the authors analyse risk in traversing stealth game levels. The authors have published a number of papers on related topics in stealth, which relate well to challenges related to vision and adversarial player movement in MOBAs.

\subsection{Patch Changes}
Like many videogames, MOBAs are frequently patched to fix bugs, rebalance game features and add new content. Because games like DOTA 2 are highly competitive, these patches also serve as a way to adjust the metagame by improving (or `buffing') heroes, items and spells which are not used often, and weakening (or `nerfing') heroes, items and spells which are overused or too powerful. Besides targeting specific entities in the game, patches can also adjust the way the game is played on a larger scale by targeting certain strategies or styles of play, adjusting the map geometry, or changing the order of drafting.

New patches are significant moments in the history of the game. A major patch changes hundreds of game elements, from the cost of an item to the number of seconds a skill is on cooldown for after use. The first matches and tournaments played professionally after a new patch typically involve a lot of speculation and experimentation, and previously dominant teams and players can suddenly find themselves scrambling to compete as new strategies are stumbled upon or hypotheses are tested.

Patches also represent a serious challenge for any AI system that relies on an understanding of the metagame (including player bot AI, commentary AI, draft analysis and more). The game fundamentally changes overnight in numerous ways, and existing archives of play data (and systems trained on them) no longer represent how the game currently plays. Two major challenges stand out in this area:

\subsubsection{Assessing Patch Impact}
In the days and weeks following a patch there is a scramble to understand what the overall effect the patch will have on the metagame. This understanding is vital for commentators to discuss in games, for teams to gain an edge on competitors, and for designers to assess the efficacy of their changes. Sometimes changes have obvious effects -- if the cost of an item increases, that item will be harder to purchase. Other effects can be subtle and may take weeks to emerge -- a buff to a particular item leads to a hero becoming stronger, thus picked more often, which in turn increases the value of a second hero whose primary purpose is to counter them. Being able to predict these chains or identify trends and strong strategies ahead of time is extremely valuable.

We believe that building a simple forward model for DOTA 2's combat may help assess the impact of small changes. Combining such a model with simple AI agents, replay data from the previous patch can be resimulated with the addition of one or more changes from the new patch, having AI agents take over at the point where the simulation diverges from the past data (for example, a patch change causes someone to live where they had died in the replay data, because the damage they took was reduced). From this, we can assess simple surface-level impacts from the patch such as changes to the efficacy of items or skills. While this is unlikely to provide a deep assessment of a patch's impact on a hero, it may provide an indication on a micro-scale about what impact a patch is likely to have.

\subsubsection{Predicting Patch Content}
Towards the end of a patch cycle the metagame often becomes stagnant -- highly-valued heroes are regularly banned and picked, many teams use the same strategies, and play approaches a state of equilibrium. A new patch will rectify this by rebalancing the game as described above, forcing strategies to be re-evaluated and a long process of experimentation and discovery to take place. A system that can assess the current metagame and predict or suggest patch changes is valuable both as a design tool (for designers, in suggesting balance changes or helping alter popular strategies) and as a competitive tool for teams to predict what new trends may emerge in the next patch, in order to prepare for them.

AI learning approaches might be able to infer potential patch changes based on trends in previous patches (heroes are often buffed or nerfed in successive patches until a particular effect is achieved). We also believe the use of a forward model, as suggested above, might be able to predict possible changes based on items, heroes and strategies which are overused or have statistically abnormal winrates. Analogies may also be made between previous patch changes and their root causes. If an item was previously rebalanced after having a high winrate or being bought too often, items following similar patterns of usage might be candidates for similar rebalancings in a future patch.

\subsection{Inventing Techniques \& Discovering Exploits}
Many of the crucial mechanical systems that are now built into DOTA 2 and many other MOBAs originated as bugs, exploits or emergent behaviour found by players in the game. For example, in DOTA 2 players earn gold by performing the killing blow on a non-hero creature. An unintentional feature of DOTA's original implementation in Warcraft 3 allowed players to attack their own creeps to kill them. Players used this feature to invent the notion of `denying', since attacking their own creeps stops their opponents from gaining gold (they received no gold themselves from doing this). This is now a fundamental part of contesting players for resources in DOTA 2, and recognised in the game through statistical tracking and its inclusion in tutorials.

Such discoveries still happen today, as patches introduce new items, skills, systems and interactions. Discovering new mechanics has benefits for many groups: it gives players and teams a temporary edge, and allows developers to fix true exploits and adjust interesting ones (denying was balanced multiple times as it became an official mechanic, while other games like League of Legends removed it entirely). An AI system that can curiously explore a MOBA's game systems to uncover new interactions and beneficial effects would be highly valued.

This challenge bears similarities to automated playtesting research \cite{zook} -- in essence, many of these interesting discovered systems can be regarded as bugs or unintentional side effects of the game's intentionally-designed systems. The special case we are considering here is whether the discovered exploits provide some kind of competitive advantage or inspire a new strategy in playing the game, perhaps similar to objective-driven mechanic discovery in platformers \cite{mechanicminer}. The most significant challenge here is in identifying progress or utility in a discovered system. Often these are highly innovative and creative in their application, which is the reason why they take so long to find despite hundreds of thousands of games taking place every day. Independently curious agents \cite{sanders} playing the game as part of automated bot matches, ignoring objectives and instead seeking novel game interactions might yield interesting results.

\subsection{Abuse}
Many MOBAs are designed to be highly competitive environments which emphasise player skill, improvement and ultimately mastery. The framing of these games as sports is further enhanced by the tight links with the professional scene in which the most popular players are put forward as role models, as well as a relentless stream of statistics and records tracking the performance of players and their ranking among their peers. This atmosphere, combined with DOTA 2's high barrier to entry and intimidating learning curve, makes the community a pressure cooker of negative emotions. Abusive players are a major problem in MOBAs.

The only attempt made to curb this behaviour is a player-run reports system which relies upon the playerbase to inform Valve when a player is acting abusively. However, this system is more often used to report players who are perceived to be playing badly, itself a form of abuse. Riot Games, developer of the MOBA League of Legends, have reportedly\footnote{http://tinyurl.com/riotresearch} applied machine learning techniques to automatically detect abusive language, although more complex problems like sarcasm, passive-aggressiveness, or abusive in-game behaviour (to intentionally sabotage another player or their own team) remain difficult to detect.

Existing research has looked into textual abuse in games \cite{languagemoba}, although much work remains to be done. The task of detecting abusive behaviour, however, including bots, scripts and intentional ability abuse is still a relatively open problem. Progress in this area benefits all games, but the richness and volume of data focused around the same structured activities makes MOBAs an appealing place to start in this case.

\section{Related Work}
Existing work investigating DOTA 2 as a domain for AI research primarily focus on the task of building AI agents to play the game. This is an extremely appealing challenge for adversarial AI research -- playing the game requires both micro-scale decision-making with fast reactions, and macro-scale strategic planning. It also has an aspect lacking in many of the recent fashionable AI challenge domains - other players. DOTA 2 is fundamentally a game about team co-operation, cohesion and communication, and this adds a much less-examined, more complex dimension to the AI problem.

The research questions which have received the most attention work with readily-available data that can be extracted from public replays, such as analyses of the heroes making up a team. In \cite{teamcomp} the authors analyse whether a more balanced distribution of roles within a team improves a team's chance of winning, a fundamental part of selecting heroes that works well together. \cite{spatiotemporal} takes a different approach, assessing the position of heroes and how they change throughout the game as an indicator of team performance (high movement and team grouping can be an indicator of success in some metagames). Another study in \cite{graphs} focuses on the relationship between hero roles and how these develop throughout different phases of the game.

Existing research of this kind may be repurposed to tackle some of the challenges outlined in this paper -- for example, understanding the phases of gameplay and the expected behaviour from different roles, as shown in \cite{graphs}, can provide a useful baseline of knowledge for commentators, provided it can change to keep up with patches and the metagame (something which, in general, existing work does not address). Similarly, research outside the AI domain but focused on MOBAs, such as the analysis of pedagogical professional streams in \cite{noob}, will help inform the creation of AI tools that teach and relay information to players and spectators.

\section{Conclusions}
In this paper we outlined a diverse range of challenges offered by the MOBA genre, using DOTA 2 as an illustration. They show that the genre has more to offer than simply a harder adversarial AI problem, with emerging challenges that pose problems for machine learning, natural language processing and generation, computational creativity, player modelling, and more. We proposed concrete projects and potential avenues for them, and discussed existing work that might relate to these challenges. MOBAs are a Frankensteinian genre, a patchwork of ideas and systems from RPGs, action games, sports, roguelikes, boardgames and more. As researchers come to tackle the problems posted by the genre, it will be crucial to pull in ideas, systems and inspiration from a wide sampling of research areas.

MOBAs are time-consuming games to learn, taking hundreds of hours to gain an average understanding of the game's main principles. One of the many ways in which this affects the game is that it makes it very difficult for researchers to come to this area as an outsider and apply their knowledge and expertise to it. Yet the genre contains within it a multitude of exciting research opportunities, vast stores of publicly available data, and professionals eager to work with new technology and ideas. Both the game's community and our research community must find ways to make these challenges amenable to people and make the genre a more accessible proposition for those who are not already experienced with it.

\bibliographystyle{aaai}
\bibliography{citations}

\end{document}